\documentclass[letterpaper, 10 pt, conference]{ieeeconf}  

\IEEEoverridecommandlockouts                              

\usepackage{graphics} 
\usepackage{epsfig} 
\usepackage{mathptmx} 
\usepackage{times} 
\usepackage{amsmath} 
\usepackage{comment}
\newcount\Comments
\usepackage{color}
\definecolor{darkgreen}{rgb}{0,0.7,0}
\definecolor{gray}{rgb}{0.95,0.95,0.95}
\definecolor{purple}{rgb}{1,0,1}

\newcommand{\kibitz}[2]{\ifnum\Comments=1{\color{#1}{#2}}\fi}

\newcommand{\sk}[1]{\kibitz{blue}{[sk: #1]}}

\newcommand{\adi}[1]{\kibitz{purple}{[adi: #1]}}

\newcommand{\map}{\mathcal{M}}

\newcommand{\VOI}{VOI\xspace}

\newcommand{\VOA}{VOA\xspace}

\newcommand{\eqVOA}{\mathcal{V}}

\newcommand{\Path}{\Pi}

\newcommand{\pathplanned}{\Path^{P}}
\newcommand{\pathplannedOne}{\Path^{P,1}}
\newcommand{\pathplannedTwo}{\Path^{P,2}}
\newcommand{\pathplannedThree}{\Path^{P,3}}
\newcommand{\pathexecuted}{\Path^{E}}

\newcommand{\loc}{t}
\newcommand{\locVar}{\tau}

\usepackage{xspace}
\usepackage{amsfonts}
\usepackage{algpseudocode}
\usepackage{subcaption}

\usepackage{url}

\title{\LARGE \bf Value of Assistance for Mobile Agents}

\author{Adi Amuzig$^{1}$, David Dovrat$^{2}$ and Sarah Keren$^{3}$
\thanks{The authors are with the Faculty of Computer Science, Technion - Israel Institute of Technology, Haifa 32000, Israel {\tt\small \{$^{1}$adi.amuzig@campus.},{\tt\small $^{2}$ddovrat@cs.},{\tt\small $^{3}$sarahk@cs.} \} {\tt\small technion.ac.il}}}%

\begin{document}

\maketitle
\thispagestyle{empty}
\pagestyle{empty}

\maketitle

\begin{abstract}
Mobile robotic agents often suffer from localization uncertainty which grows with time and with the agents' movement. This can hinder their ability to accomplish their task.
In some settings, it may be possible to perform assistive actions that reduce uncertainty about a
robot’s location.
For example, in a collaborative multi-robot system, a wheeled robot can request assistance from a drone that can fly to its estimated location and reveal its exact location on the map or accompany it to its intended location.
Since assistance may be costly and limited, and may be requested by different members of a team,
there is a need for principled ways to support the decision of which assistance to provide to an agent and when, as well as to decide which agent to help within a team.
For this purpose, we propose {\em Value of Assistance (\VOA)} to represent the expected cost reduction that assistance will yield at a given point of execution. 
We offer ways to compute \VOA based on estimations of the robot's future uncertainty, modeled as a Gaussian process.
We specify conditions under which our \VOA measures are valid and empirically demonstrate the ability of our measures to predict the agent's average cost reduction when receiving assistance in both simulated and real-world robotic settings.
\end{abstract}

\section{Introduction}
In many real-world settings, autonomous agents suffer from various forms of uncertainty which may hinder their ability to accomplish their assigned tasks. 
In some settings, it may be possible to perform actions to reduce this uncertainty. These actions can be performed by the agent itself (e.g. by activating a sensor) or by another agent (e.g. by communicating localization information).
Since such actions may be costly and the opportunity to perform them may be scarce, 
there is a need for principled measures to assess their expected benefit and to find the best time to execute them. 
Moreover, when assistance can be given to only one member of a team, such unified measures can support the decision of which agent to assist. 

For this purpose, we propose {\em Value of Assistance} (\VOA) to represent the cost reduction an assistive action is expected to yield. 
\VOA~is inspired by the well-established notion of {\em Value of Information} \cite{4082064,RUSSELL1991361,ZLtr9635}, which is used to quantify the impact information has on autonomous agents' decisions. We offer \VOA~to quantify the expected impact assistance, in the form of physical assistance or information provision, would have on the navigating mobile agents' ability to achieve their tasks. 

This idea can be instantiated in different ways for various embodied agents with different characteristics. Here, we consider mobile robotic agents that are tasked with following a path, defined as a sequence of waypoints that need to be visited.
The agents suffer from localization uncertainty that grows as they progress in the environment.
This uncertainty can be characterized as a Gaussian process, where each coordinate of the agent's location is a normally distributed random variable with growing variance.
This is particularly relevant in settings in which robots 
rely only on internal sensors for localization rather than on sensing the external environment. 

\begin{figure}[t]
    \centering
    \includegraphics[width=7.5cm]{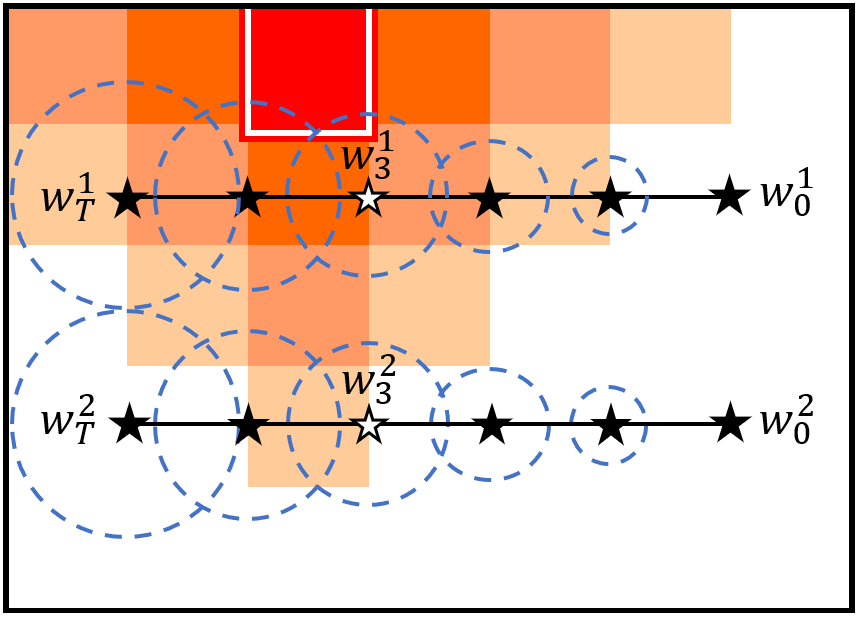}
    \caption{Demonstrating \VOA~for a team of two mobile robots.}
    \label{fig: intro example}
\end{figure}

To demonstrate \VOA, consider the setting depicted in Figure \ref{fig: intro example} in which two mobile robots navigate specified paths, each defined as a sequence of waypoints $w^i_0\dots w^i_T$. 
Both agents start with full certainty about their initial location $w^i_0$, but accumulate localization uncertainty according to the same model as they move along the intended path (uncertainty is depicted by the dashed circles). 

Consider a drone that can fly to one of the agents and provide localization information at a planned waypoint, but can do so only once.
This raises the need to weigh the benefit of 
providing assistance at waypoint $w^i_3$ to either agent. On the one hand, if the robots can use known landmarks on the map to improve localization, robot $1$ can use the mapped obstacle on the top (white border) to improve its localization accuracy, reducing its need for help. If, 
however, the robots do not have the ability to sense the presence of the obstacle, robot $1$ is in greater need of help due to its proximity to it. 



The example above demonstrates how \VOA~depends on the agents' capabilities and tasks, as well as on the available assistance and the environment.
Beyond this illustrative scenario, 
\VOA is a relevant metric for various multi-agent and single-agent applications.
For example,  
in multi-agent underwater AUV navigation\cite{chandrasekhar2006localization,cario2021accurate}, agents cannot use a GPS for localization while submerged but can rise to the surface to acquire their exact location. Similar considerations occur in search and rescue missions \cite{9345802,Koch2016-lc}, indoor navigation \cite{8692423}, and more.

We present methods to assess \VOA~for navigating mobile agents for two forms of assistance:  
{\em Relocation} corresponds to settings in which available assistance consists of placing an agent at its intended location or accompanying it to that location (after which the agent knows its exact location);
{\em Localization} is relevant to settings in which it is only possible to reveal to an agent its exact current location, but the agent needs to navigate on its own to its intended location after the intervention. 

We make the following contributions.
\begin{enumerate}
    \item We introduce the concept of {\em Value of Assistance} (\VOA) for mobile autonomous agents.
    \item We formalize \VOA for two different forms of assistance: relocation and localization. 
    \item We empirically demonstrate the ability of \VOA to predict where assistance will be most beneficial in both simulation and real-world robotic settings.
\end{enumerate}

\section{Related Work}
Choosing when to request assistance and when to acquire new information
is a fundamental problem in a variety of single-agent and multi-agent AI scenarios 
\cite{wooldridge2009introduction}. 
Within this variety, many frameworks focus on information-sharing settings, in which information acquisition can be considered either implicitly within the agent's action space \cite{10.1145/860575.860598} or as an explicit observation or communication action \cite{10.1007/10720246_21,Marcotte2020-wo}.
Our focus is on the latter case, and on providing measures for assessing the benefit of performing such actions. 

One way to reason about the effect information will have on performance is via {\em Value of Information} (\VOI) \cite{4082064,RUSSELL1991361,stratonovich2020theory}. The basic notion of \VOI represents the expected increase in an agent's value, or reduction in cost, as a result of obtaining the information. This idea has been extended in various ways \cite{HANSEN2001139,ZLtr9635,Chen_Javdani_Karbasi_Bagnell_Srinivasa_Krause_2015,Chen_Choi_Darwiche_2015,GZaaai97ws,DBLP:journals/corr/abs-1302-3596,10.5555/2484920.2485011}. Recently,
\VOI~was adapted to a human-robot collaborative setting in which {\em Value of Helpfulness} is used to quantify the extent by which an agent's actions can reduce the cost of a human's plan  \cite{https://doi.org/10.48550/arxiv.2010.04914}.
Another line of work uses {\em Value of Communication} to assess the expected benefit of exchanging information within a team \cite{https://doi.org/10.1111/j.1467-8640.2008.01329.x,ijcai2020p36}. 
Our \VOA~measure 
adapts \VOI~to consider the expected benefit of providing assistance to mobile agents for which localization uncertainty can be described as a Gaussian process. 
Our use of a Gaussian process to model the localization uncertainty in our \VOA~computation is inspired by similar uses in quantifying collision probability \cite{4795553,5738354} and in
selecting a path with minimal localization uncertainty by gathering information from landmarks along the path \cite{prentice2009belief,van2012motion}.

\begin{figure*}[ht]
\vspace{0.18cm}
\centering
\begin{subfigure}{0.48\textwidth}
\centering
    \includegraphics[width=0.8\textwidth]{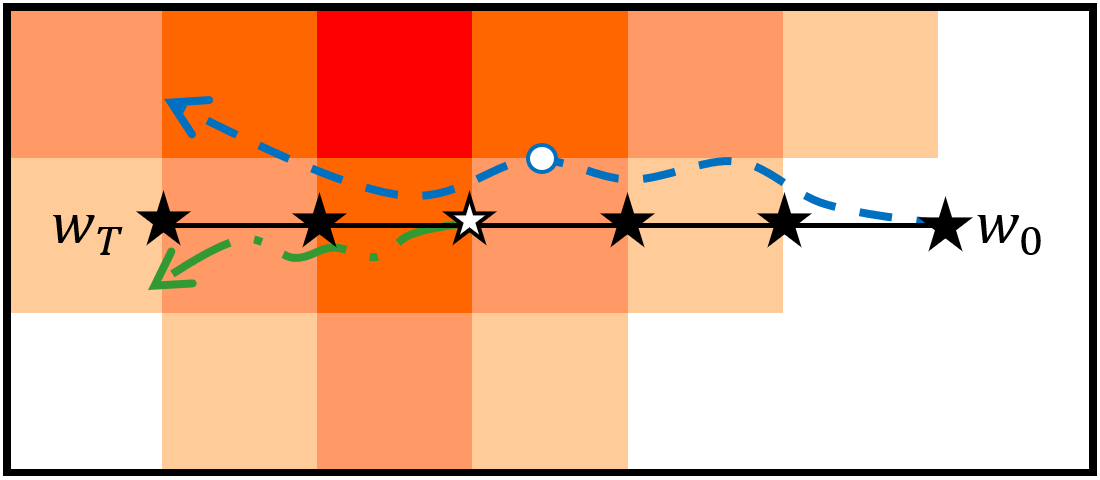}
        \caption{}
        \label{fig: validation relocation}
    \end{subfigure}~
    \begin{subfigure}{0.48\textwidth}
    \centering
    \includegraphics[width=0.8\textwidth]{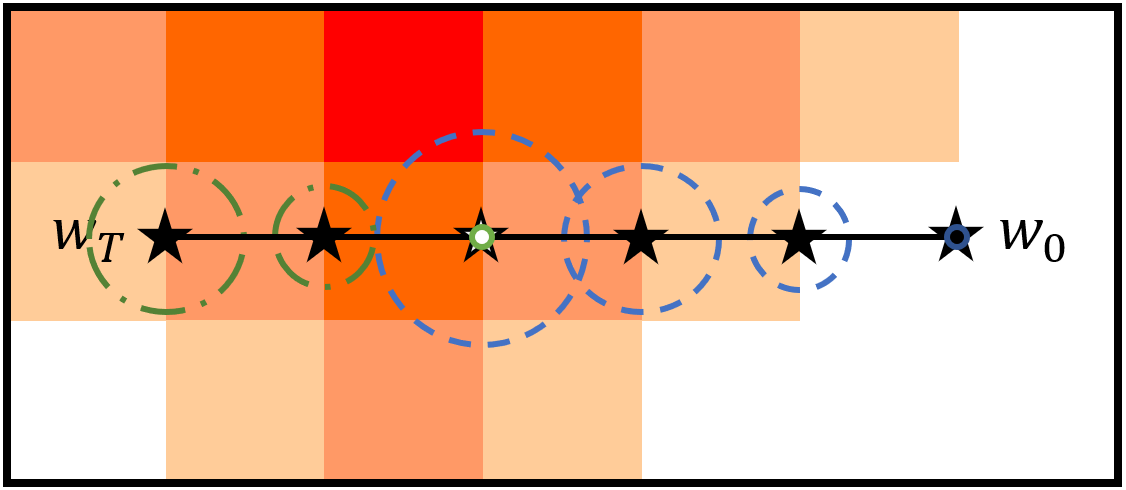}
        \caption{}
        \label{fig:relocation-u}
    \end{subfigure}

    \begin{subfigure}{0.48\textwidth}
    \centering
    \includegraphics[width=0.8\textwidth]{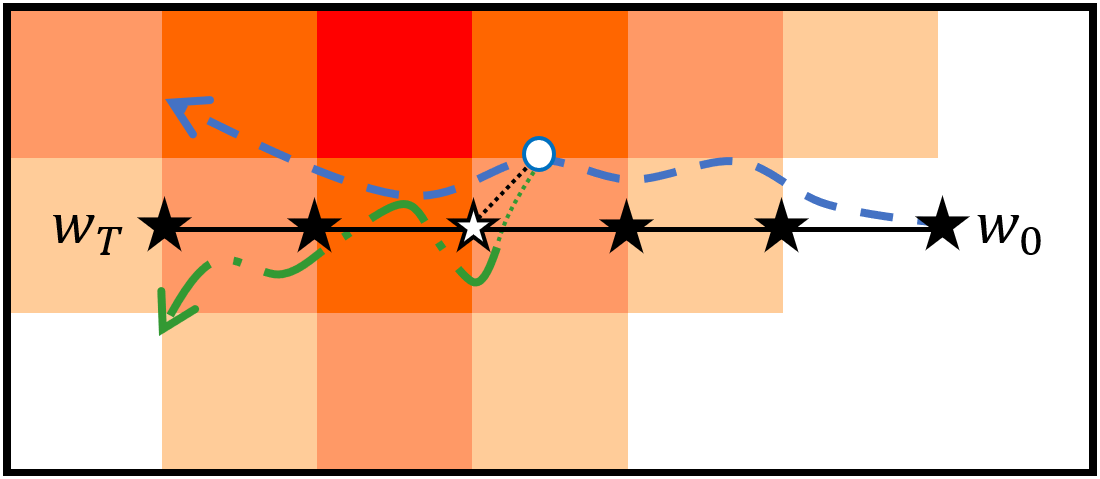}
        \caption{}
        \label{fig: validation localization}
    \end{subfigure}~
    \begin{subfigure}{0.48\textwidth}
    \centering
    \includegraphics[width=0.8\textwidth]{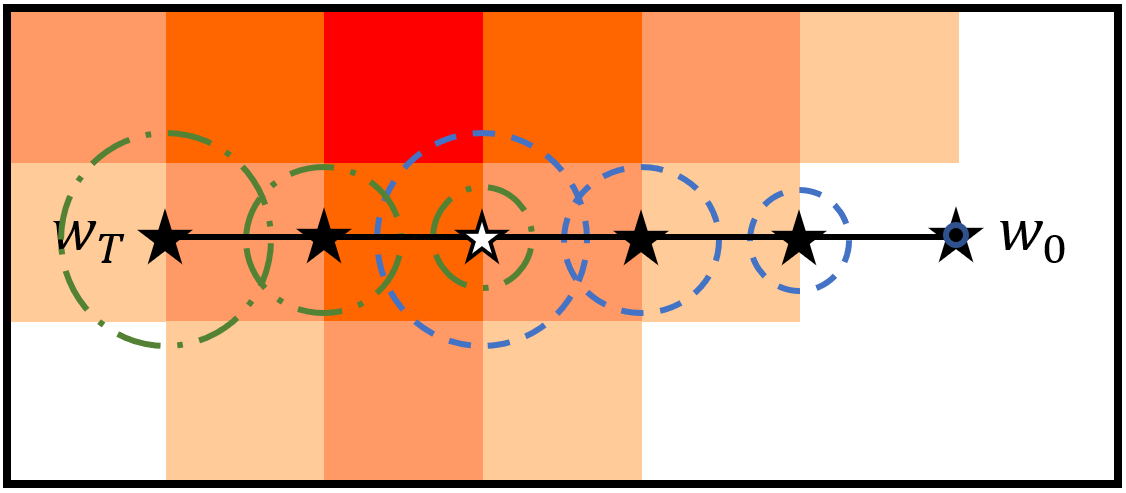}
        \caption{}
        \label{fig:localization-u}
    \end{subfigure}
\caption{ 
Demonstrating relocation assistance (top) and localization assistance (bottom). } 

    \label{fig:example-cont}
\end{figure*}
    
\section{Background}\label{sec:bg}

Localization uncertainty of mobile agents may be caused by movement error or unmodeled system dynamics. 
As is common in many related frameworks, we model an agent's location over time as a Gaussian process \cite{AFSHARI2017218,9610137}. 
Thus, an agent's location $x$ at time $t$ is described as a Gaussian distribution $
    {x}\left({t}\right) \sim \mathcal{N}\left({\hat{x}\left({t}\right), \Sigma_{x\left({t}\right)}}\right)
$
where $\hat{x}\left({t}\right)$ is the mean location and $\Sigma_{x\left({t}\right)}$ is the covariance of the location vector $x$ at time $t$.

We model the agent's kinematics as
\begin{equation}\label{eq: sys_dynamics}
    \dot{x}\left({t}\right)
    =
    f\left({{x}\left({t}\right), u\left({t}\right)}\right)
\end{equation}
where $u\left({t}\right)$ is the movement reference or command signal, which reflects the autonomous agent's desired change of location at time $t$.
Accordingly, we model the agent's location \emph{covariance function} as
\begin{equation}\label{eq: sys_covariance}
    \frac{d}{dt}\Sigma_{x\left({t}\right)} = g\left({\Sigma_{x\left({t}\right)}, u\left({t}\right)}\right).
\end{equation}


    
 \section{Value of Assistance (\VOA)} \label{sec: VOA}

We define \emph{Value of Assistance} (\VOA) as the difference between the expected cost of following a planned path and the expected cost of following the same path, but when receiving assistance.
We first present a method for computing the expected cost of a path for a navigating robot with localization uncertainty and then offer ways to compute \VOA using two forms of assistance: \emph{relocation}, and \emph{localization}.  

Both forms of assistance are demonstrated in Figure \ref{fig:example-cont}. Figures \ref{fig: validation relocation} and \ref{fig: validation localization} show the agent's progress on its planned path from $w_0$  to $w_T$ for relocation and localization assistance, respectively. Without assistance, the agent follows the path depicted by the blue dashed line. In Figure \ref{fig: validation relocation}, the agent receives relocation assistance at $w_3$, and is moved from its actual location (white dot) to its intended location (white star) and thereon follows the path depicted by the green dash-dot line.
In Figure \ref{fig: validation localization}, the agent receives localization assistance at $w_3$, and knows its true location (white dot) at the time of intervention.
The agent tries to return to its intended location (white star) by following a {\em corrective path}, depicted by the dotted black line.
However, it actually follows the finely dotted green line while it
accumulates additional localization error as it moves.
Once the agent believes it reached its intended location, it continues to execute its planned path from its actual location, resulting in a motion represented by the green dash-dot line.

Figures \ref{fig:relocation-u} and \ref{fig:localization-u} depict the progression of the estimated location uncertainty for relocation and localization, respectively.
The evolution of the localization uncertainty before the intervention is marked by the blue dashed circles. Uncertainty after the intervention, including uncertainty accumulated over the corrective path for localization, is marked by the dash-dot green circles. 

We note that our current account only considers the expected benefit of providing assistance at a given point of execution, abstracting, for now, the optimization considerations of the helping agent. Thus, we currently assume that the possible interventions are given as input, e.g. by specifying the achievable range of the helping agent due to its fuel constraint. Our focus is on supporting the decision of when to perform an intervention and selecting the agent that should be assisted within a team. 
Furthermore, the proposed formulations are for a dynamics model in which $g$ in equation \ref{eq: sys_covariance} is equal to $\Sigma_{x\left({t}\right)} + g'\left({u\left({t}\right)}\right)$ where $g'$ represents the increase in location uncertainty due to the movements of the agent.
Optimizing the behavior of the helping agent,
and expanding the formulation for more complex models 
are important avenues for future work. 

\subsection{Cost of Following a Path}
\label{sec: Expected Cost of Following a Path}
We use $\locVar \in \left[{0,1}\right]$ to denote the progress on a path $\Path$, such that
 $\Path(\locVar)$ represents a point on the path.
 For example, $\Path\left({0}\right)$ is at the beginning of the path, $\Path\left({1}\right)$ is at the end, and $\Path\left({\frac{1}{2}}\right)$ is half way through.
 If we consider an agent that moves along the path at a constant speed, we can think of $\locVar$ either in terms of time or distance without ambiguity. In any case, $\locVar$ represents the ratio between elapsed and total time (or distance) on the path. 

We aim to compute \VOA~by considering the cost difference that can be achieved via assistance in settings in which a cost (or reward) is associated with traversing specific locations. 
There are different ways to assess the cost of following a given path.
Here, we assume a user-defined cost-map is used to represent any domain and application-specific considerations including both costs and rewards. For example, it can be used to represent high-risk areas the agent should avoid and high-reward areas it should traverse. 

Given a map, we rate the cost of following path $\Path$, represented as a continuous function
$\Path : \left[{0,1}\right] \rightarrow \mathbb{R}^{n}$,
by summing all points of the path on a given semi-continuous cost-map $\map : \mathbb{R}^{n} \rightarrow [0,1]$.


Consider a planned path $\pathplanned$, and a path $\pathexecuted$ actually executed.
If the agent is capable of perfectly following the planned path, then $\pathexecuted\left({\locVar}\right) \equiv \pathplanned\left({\locVar}\right)$ $\forall \locVar$. 
However, if the agent is unable to follow the exact planned path due to localization uncertainty, we need a way to model this uncertainty in order to reason about the actual outcome of its movements.
We model this uncertainty using  a Gaussian process with a known covariance function $\Sigma : \mathbb{R}^{n} \times \left[{0,1}\right] \rightarrow \mathbb{R}^{n \times n }$
 for each point $\pathplanned\left({\locVar}\right)$ on path $\pathplanned$.
We assume that without extraneous intervention, the agent's uncertainty grows monotonically, i.e., each element in $\Sigma\left(\locVar_{1}\right)$ is not greater than its corresponding element in $\Sigma\left(\locVar_{2}\right)$ if $\locVar_1 < \locVar_2$.

Let $X\left({\locVar}\right)$ denote the random variable describing all possible outcomes of $\pathexecuted\left({\locVar}\right)$, s.t.
\begin{equation}
\label{eq: Gaussian location uncertainty}
    X\left({\locVar}\right)\sim  \mathcal{N}\left({\pathplanned\left({\locVar}\right), \Sigma\left({\pathplanned\left({\locVar}\right), \locVar}\right)}\right).
\end{equation}

The expected accumulated cost for an agent to follow path $\pathplanned$ from $\locVar_0$ to $\locVar_1$ on cost-map $\map$ 
 can be computed as follows.  
\begin{equation}
\label{eq:pathPortionCost}
C_{\locVar_0}^{\locVar_1}\left({\pathplanned}\right)
=
\int_{\locVar_0}^{\locVar_1} \int_{\mathbb{R}^n} P\left({X\left({\loc}\right)
= \boldsymbol{x}}\right) \map\left({\boldsymbol{x}}\right) d\boldsymbol{x} d\loc,
\end{equation}
and the total expected cost for following the entire path $\pathplanned$ from $0$ to $1$ is 
\begin{equation}\label{eq:pathcostl}
C\left({\pathplanned}\right) = C_{0}^{1}\left({\pathplanned}\right).
\end{equation}

\subsection{Computing Value of Assistance (\VOA)}
We aim to provide measures that can assess the expected benefit of an assistive action in terms of its effect on the expected cost. We denote \VOA~for path $\Path$ on map $\map$ as $\eqVOA \left({\map,\Path,\locVar}\right)$, where the 
agent receives assistance at point $\locVar$. 
Since the expected cost $C_{0}^{\locVar}\left({\pathplanned}\right)$ of following a planned path $\pathplanned$ up to $\locVar$ is the same with and without intervention, we only need to assess the difference in the expected cost of following the remainder of the path.
We denote the planned path after the intervention at $\locVar$ as $\pathplanned_\locVar$. The corresponding costs for following the remainder of the path without and with intervention are denoted as 
$C_{\locVar}^{1}\left({\pathplanned}\right)$ and $C_{\locVar}^{1}\left({\pathplanned_{\locVar}}\right)$, respectively.
This path $\pathplanned_{\locVar}$ depends on the nature of the assistance provided to the agent. 


 

\subsubsection{Relocation Assistance}
After attempting to follow the planned path up to the point of intervention $\locVar$, an agent receives {\em relocation assistance} 
 and is moved from its actual location, denoted $\pathexecuted(\locVar)$, to the corresponding point on the planned path, $\pathplanned(\locVar)$.
Depending on the application, this can correspond to physically transporting the agent to $\pathplanned(\locVar)$ or accompanying it to that location. 
In any case, when the agent is left at $\pathplanned(\locVar)$ to continue performing the planned path, it has full certainty about its location.

Denote $X'(\loc)$ as the random variable describing the possible location of $\pathexecuted(\locVar)$ after the intervention at $\locVar$ , i.e. $\forall$ $\loc$ s.t. $\locVar < \loc \leq 1$,
\begin{equation}
\label{eq: Relocation uncertainty}
    X'\left({\loc}\right) \sim \mathcal{N}\left({\pathplanned\left({\loc}\right),
    \Sigma'\left({\pathplanned\left({\loc}\right), \loc}\right) }\right)
\end{equation}
where
\begin{equation}
\label{eq: Relocation variance}
\Sigma'\left({\pathplanned\left({\loc}\right), \loc}\right) = \Sigma\left({\pathplanned\left({\loc}\right), \loc}\right) - \Sigma({\pathplanned(\locVar), \locVar}).
\end{equation}
In Figure \ref{fig:relocation-u}, the green dash-dot circles show the re-initialization of the probability distribution covariance at $\locVar$ and the evolution of $\Sigma'$ over time after the intervention.

The expected cost of the agent's attempt to follow path $\pathplanned$ after the intervention for $\locVar < \loc \leq 1$ is computed as
\begin{equation}
\label{eq: estimated grade of path with relocation}
C_{\locVar}^1\left({\pathplanned_{\locVar}}\right)
=
\int_{\locVar}^1
\int_{\mathbb{R}^n} P(X'\left({\loc}\right) = \boldsymbol{x}) \map(\boldsymbol{x}) d\boldsymbol{x} d\loc.
\end{equation}

The total expected cost of following the entire path with the intervention at $\locVar$ becomes
\begin{equation}
\label{eq:voa-relocation}
C\left({\pathplanned_{\locVar}}\right)
=
C_0^{\locVar}\left({\pathplanned}\right)
+
C_{\locVar}^1\left({\pathplanned_{\locVar}}\right).
\end{equation}

The value of assistance can therefore be calculated as
\begin{equation}\label{eq:voa-reloc}
\eqVOA ({\map,\pathplanned,\locVar})
=
C_{\locVar}^1\left({\pathplanned}\right)
-
C_{\locVar}^1\left({\pathplanned_{\locVar}}\right).
\end{equation}

\subsubsection{Localization Assistance} \label{sec: localization assistance}
 
Another form of assistance is to inform the agent of its true location at point $\pathexecuted\left({\locVar}\right)$. After this intervention, the agent attempts to return on its own to its intended location $\pathplanned\left({\locVar}\right)$. While it has full localization certainty at the time of intervention, it accumulates uncertainty while performing this corrective behavior, referred to as the {\em corrective plan}. 
The accumulated uncertainty from the point after the agent finished the correction and its progression thereon is illustrated in Figure \ref{fig:localization-u} (green dash-dot circles).

We divide the planned paths for this scenario into three distinct sections. 
The first is from the beginning of the original planned path $\pathplanned(0)$ to the point of intervention $\pathplanned(\locVar)$,
marked as $\pathplannedOne$. 
The second is the corrective plan from $\pathexecuted(\locVar)$ to $\pathplanned(\locVar)$, denoted as $\pathplannedTwo$.
The third corresponds to the agent's plan
for the remainder of $\pathplanned$,
from $\pathplanned(\locVar)$ to $\pathplanned(1)$,
denoted as $\pathplannedThree$.

To compute $\pathplannedTwo$, consider a corrective path $\pathplannedTwo_i$ from the location at the time of intervention $\boldsymbol{x}_i \in \mathbb{R}^n$ to the intended location $\pathplanned\left({\locVar}\right)$.
Denote $X_{i}(\loc)$ as the random variable describing the possible locations of an agent attempting to follow the corrective path $\pathplannedTwo_i$, i.e. $\forall \loc$ s.t. $0 \leq \loc \leq 1$,
\begin{equation} \label{eq: Relocation general rand val}
X_i\left({\loc}\right)
\sim
\mathcal{N}\left({\pathplannedTwo_i\left({\loc}\right), \Sigma_i\left({\pathplannedTwo_i\left({\loc}\right), \loc}\right)}\right).
\end{equation}

The total expected cost for an agent to follow $\pathplannedTwo_i$ is
\begin{equation}
\label{eq: return path cost}
C\left({\pathplannedTwo_i}\right)
=
\int_{0}^1
\int_{\mathbb{R}^n} P(X_i\left({\loc}\right) = \boldsymbol{x}) \map(\boldsymbol{x}) d\boldsymbol{x} d\loc.
\end{equation}

Conditions under which the expected cost of the corrective path can be computed as a Gaussian process are described in the supplementary material\footnote{Our codebase, all our results, and our complete formal account are provided in the supplementary material: 
\url{https://github.com/CLAIR-LAB-TECHNION/VOA}}.
Under such conditions, the cost can be computed according to the probability of the agent's location at the point of intervention,

\begin{equation}
\label{eq: total estimated grade of correction path}
C\left({\pathplannedTwo}\right)
=
\int_{\mathbb{R}^n}
P\left({X\left({\locVar}\right) = \boldsymbol{x}_i}\right)
C\left({\pathplannedTwo_i}\right)
d\boldsymbol{x}_i.
\end{equation}

The expected localization uncertainty after executing the corrective path denoted as $\pathplannedTwo\left({1}\right)$, is 
\begin{equation}
\label{eq: total estimated uncertainty of correction path}
 \small   
\Sigma^2\left({\pathplannedTwo\left({1}\right), 1}\right)
=
\int_{\mathbb{R}^n}
P\left({X\left({\locVar}\right) = \boldsymbol{x}_i}\right)
\Sigma_i\left({\pathplannedTwo_i\left({1}\right),1}\right) d\boldsymbol{x}_i.
\end{equation}

After completing the corrective path $\pathplannedTwo$, the remainder of the path $\pathplannedThree$ 
can be described for $\locVar < \loc \leq 1$ by the random variable
\begin{equation}
\label{eq: After Relocation uncertainty}
X^3\left({\loc}\right)
\sim
\mathcal{N}\left({\pathplanned\left({\loc}\right),
\Sigma^3\left({\pathplanned\left({\loc}\right), \loc}\right)
}\right),
\end{equation}
where the location uncertainty is
\begin{equation}
\label{eq: after relocation variance}
\begin{array}{l}
     \Sigma^3\left({\pathplanned\left({\loc}\right), \loc}\right) =
     \\
     \Sigma\left({\pathplanned\left({\loc}\right), \loc}\right)
 +
 \Sigma^2\left({\pathplannedTwo\left({1}\right), 1}\right)
 -
 \Sigma\left({\pathplanned\left({\locVar}\right), \locVar}\right).
\end{array}
\end{equation}

The expected cost of following $\pathplannedThree$ is computed as
\begin{equation}
\label{eq: estimated grade of path after locatlization}
C\left({\pathplannedThree}\right)
= 
\int_{\locVar}^1
\int_{\mathbb{R}^n}
P(X^3\left({\loc}\right) = \boldsymbol{x})
\map\left({\boldsymbol{x}}\right)
d\boldsymbol{x} d\loc.
\end{equation}

The total expected cost of executing $\pathplanned_{\tau}$ becomes
\begin{equation}
C\left({\pathplanned_{\locVar}}\right)
=
C_0^{\locVar}\left({\pathplannedOne}\right)
+
\left({
C\left({\pathplannedTwo}\right)
+
C\left({\pathplannedThree}\right)
}\right).
\end{equation}

The calculation of the total value of assistance is therefore 
\begin{equation}
\label{eq: voa localization}
\eqVOA\left({\map, \pathplanned, \locVar}\right)
=
C_{\locVar}^1\left({\pathplanned}\right)
-
\left({
C(\pathplannedTwo)
+
C(\pathplannedThree)
}\right).
\end{equation}

\section{Empirical Evaluation}\label{sec: Empirical Evaluation}
The objective of our evaluation is to examine the ability of our proposed \VOA~measures to predict the effect relocation and localization assistance will have on an agent's performance in both simulated and real-world robotic settings. 
For this purpose, we compare \VOA to the \emph{Average Executed Cost Difference} (AECD) achieved in our experiments. AECD represents the empirical difference between the average cost of executing a planned path without assistance and the average cost when assistance is provided at a given point.
\begin{figure}
    \centering
    \begin{subfigure}{0.48\textwidth}
        \includegraphics[width=0.95\textwidth]{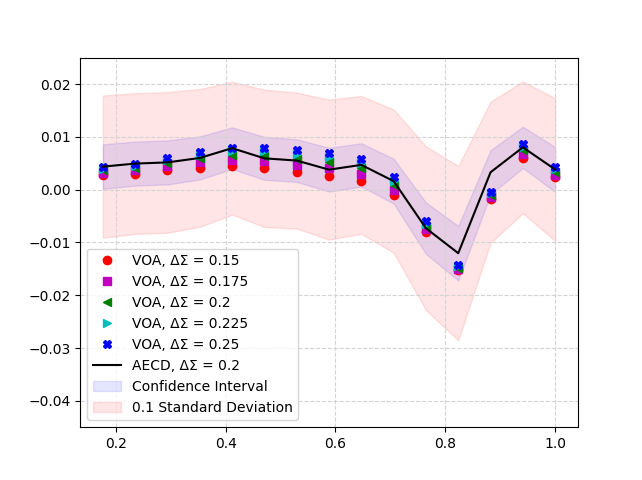}
        \caption{Relocation assistance}
        \label{fig: Relocation VOA to validation}
    \end{subfigure}
    \hfill
    \begin{subfigure}{0.48\textwidth}
        \includegraphics[width=0.95\textwidth]{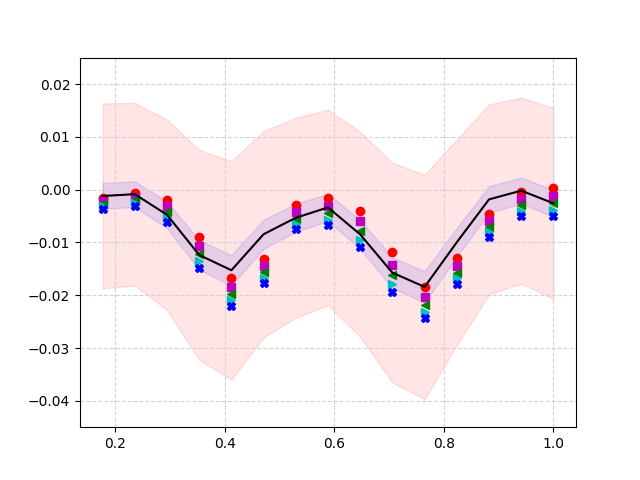}
        \caption{Localization assistance}
        \label{fig: localization voa to validation}
    \end{subfigure}
    \caption{AECD and VOA (y-axis) per path progression $\locVar$ (x-axis) for a map-path pair. AECD is plotted with its 99\% confidence interval and 0.1 standard deviations.}
    \end{figure}
    

\subsection{Simulation}\label{sec: simulation}
We generated a simulation
of a navigating agent according to the settings illustrated in Figure \ref{fig:example-cont}. To support various environments, we randomly generated cost-maps as $100 \times 100$ grids specifying the cost /\ reward of traversing each cell of size $1 \times 1 [m]$. To simulate the agent's planned path $\pathplanned$, we generated a straight path on each map.
On each path, we sampled $15$ different waypoints at 
which administering assistance to the navigating agent is possible.

Each simulation started with the agent at the beginning of the path, located at waypoint $\boldsymbol{w}_0$, 
with a fixed velocity $\left\Vert{\boldsymbol{v}}\right\Vert = 1[m/s]$.
We set the number of simulation steps to be $m$, the natural number closest to the planned path's length.  
Each step performed by the agent was calculated by adding the velocity vector to the agent's location and then adding another vector, determined by a random variable normally distributed with zero mean and
$\Delta \Sigma I_2 $
as its covariance matrix, where $I_2$ is the identity matrix, and $\Delta \Sigma = 0.2$.
Consequently, the increase in location uncertainty 
is linked to the distance covered by the agent along the path. 
For each simulation, the path cost without assistance was computed according to Equation \ref{eq:pathcostl} by summing the values of the traversed cost-map cells according to the coordinates at each of the $m$ steps. 

\subsubsection{Relocation Assistance}
To test our \VOA~measure for relocation (Equation \ref{eq:voa-reloc}), 
we summed the agent's first $l$ steps, and then let it continue an additional $m-l$ steps to record a single unassisted run.
We then placed a simulated agent at $\boldsymbol{w}_0 + l\boldsymbol{v}$, corresponding to relocating it to $\pathplanned(\locVar)$, meaning that the agent received assistance at $\locVar = l/m$.
After the intervention, the agent completes its remaining $m-l$ steps.

We performed $1000$ simulation runs in which the agent received assistance at a certain waypoint for each of the $15$ candidate waypoints of each path.
For every simulation in which the agent was assisted at some waypoint, we ran an unassisted run, that differed only from the point of intervention onward.
We calculated AECD as the difference between the average cost with and without assistance over all these simulation runs.
Repeating these calculations for 100 randomly generated maps, 
we ran a total of 3 million simulated scenarios.

We numerically computed \VOA~for each setting by using a Gaussian kernel
$\mathcal{K}\left({\mu, \Sigma}\right)$ represented as a $100 \times 100$ matrix.
According to Equation \ref{eq: Relocation variance},
if the assistance point was at distance $l$ from $\boldsymbol{w}_0$ and $l \leq k \leq m$,
\begin{equation}
\label{eq: kappa relocation}
\mathcal{K}^{k} =  \mathcal{K}\left({ \boldsymbol{w}_k,  k \Delta \Sigma I_2}\right)
-
\mathcal{K}\left({\boldsymbol{w}_k,  \left({k-l}\right) \Delta \Sigma I_2}\right),
\end{equation}
where the uncertainty increase rate is  $\Delta \Sigma$, 
and $\boldsymbol{w}_k = \boldsymbol{w}_0 + k \boldsymbol{v}$.
According to Equation \ref{eq:voa-reloc},

\begin{equation}
\label{eq: empirical voa relocation}
\eqVOA\left({\map,\pathplanned,\locVar = l/m}\right)
= 
\sum_{k=l}^m
\sum_{i=1}^{100}
\sum_{j=1}^{100}
\mathcal{K}^{k}_{i,j}
\mathcal{M}^{k}_{i,j},
\end{equation}
where
$\mathcal{M}^{k}$ is a 
$100 \times 100$ matrix copy of the map centered at $\boldsymbol{w}_k$,
and $i$ and $j$ are the row and column indices of the matrices.

\noindent{\bf Results:} Figure \ref{fig: Relocation VOA to validation} shows the VOA values calculated for a single map-path pair with different $\Delta\Sigma$ values on top of the AECD we attained from the analysis of the simulations described above per path progression $\locVar$ (results for additional map-path pairs can be found in the supplementary material).

Results show that \VOA values are similar to AECD, even with up to $25\%$ difference between the $\Delta\Sigma$ used in the simulation and the one used in the \VOA calculation. This indicates that \VOA provides a reliable estimation of the actual value of providing assistance, without being sensitive to miss-estimations of the variance of the movement error.

\begin{figure}[t!]
    \centering
        \includegraphics[width=0.49
        \textwidth]{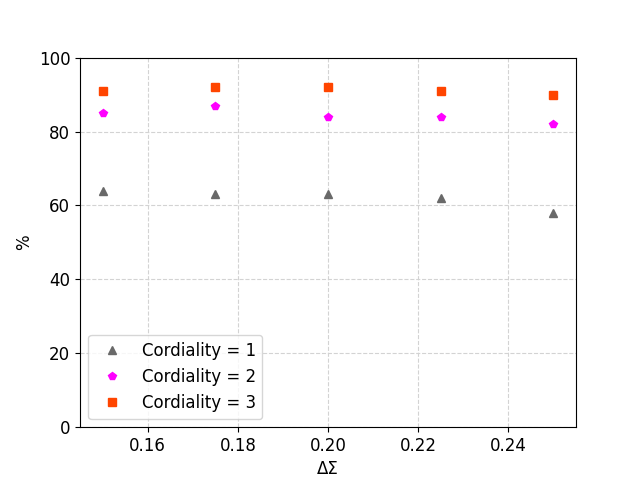}
        \label{fig: Relocation correct multi max}
    \caption{Relocation assistance: percentage of map-path pairs in which the waypoint with maximal AECD was within the sets of $1$, $2$, and $3$ waypoints with maximal VOA values.}
    \label{fig: correct multi max}
\end{figure}

After validating that \VOA predicts the effect of an assistive action, 
we examine whether it 
indicates the waypoint at which it would be best to intervene.
For this, we present in Figure \ref{fig: correct multi max} 
the percentage of map-path pairs for which the AECD value was within the $1$, $2$, or $3$ maximal \VOA values for different $\Delta\Sigma$ values. 
Results show that the maximal AECD value is within the top $1$,$2$, and $3$ maximal \VOA values for $60\%$, $85\%$, and $92\%$ of the instances on average, respectively.

As can be seen in Figure \ref{fig: Relocation VOA to validation}, 
there are a few waypoints with an AECD value close to the maximal AECD.
This goes to show that choosing any one of the top 3 \VOA~valued waypoints will achieve a result similar to choosing the waypoint with the maximal AECD
(in the supplementary material we provide an in-depth account of the magnitude of the value differences).
 






\subsubsection{Localization Assistance}\label{sec: simulation localization}

The difference between AECD for localization and relocation is that AECD for localization needs to account for the corrective path. 
We set the agent's velocity at the point of intervention $\pathexecuted(\locVar=l/m)$ to be  
\begin{equation}
\label{eq: temp vel}
\boldsymbol{u} = \frac{\pathplanned\left(\locVar \right) - \pathexecuted\left(\locVar\right)}{\left\Vert{\pathplanned\left(\locVar\right) - \pathexecuted\left(\locVar\right)}\right\Vert}
\end{equation}
for $m'$ steps, where $m'$ is the natural number closest to  $\left\Vert{\pathplanned\left(\locVar\right) - \pathexecuted\left(\locVar\right)}\right\Vert$,
until changing the velocity back to $\boldsymbol{v}$ for the completion of the path.

We performed $5000$ simulations in which the agent received assistance at a candidate waypoint for each of the $15$ waypoints of each path. As we did for relocation, we paired each of these runs with an unassisted run.
We calculated AECD as the difference between the average cost with and without assistance, for 100 randomly generated maps, 
resulting in a total of 15 million simulated scenarios.
We calculated \VOA in a manner similar to the relocation experiment, except that here we added the cost of the corrective path $\pathplannedTwo$. 
The details of the full numeric computation are provided in the supplementary material and omitted here for brevity.

\noindent{\bf Results:} 
Figure \ref{fig: localization voa to validation} shows that the \VOA values are close to AECD,
as was the case for relocation assistance.
In the supplementary material 
we show 
that the results of the cordiality analysis are also comparable.

\begin{figure}[t!]
\vspace{0.18cm}
\centering 
    \includegraphics[width=0.48\textwidth]
    {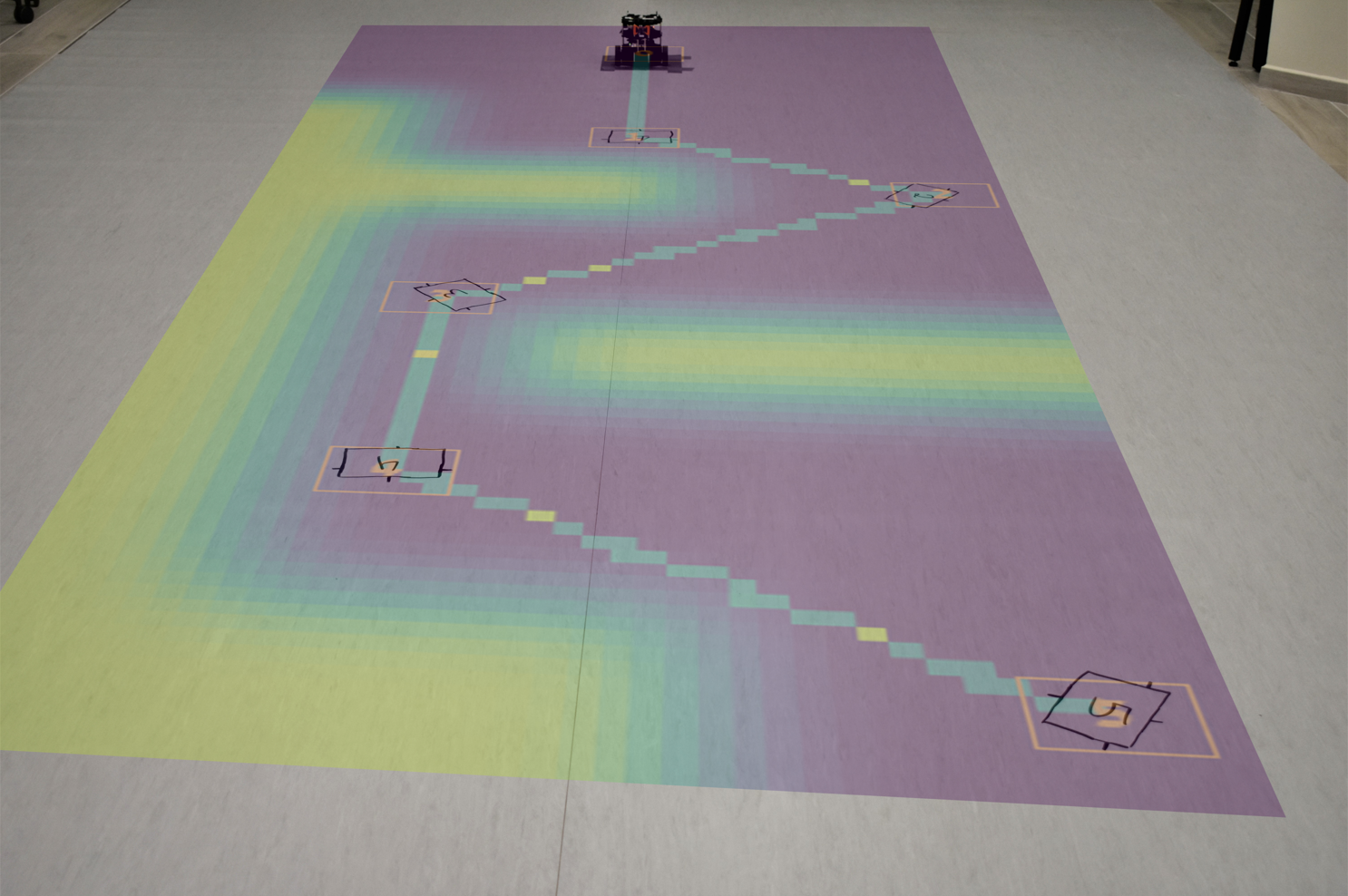}
        \caption{The experimental setup.
        The 6 waypoint path and a cost-map are projected on the laboratory floor.
        Our Turtblebot3 is at the first waypoint.
        }
\label{fig: Turtlebot full path}
\end{figure}
    
\subsection{Lab Experiments}

To test \VOA~in a real-world setting, we employed a mobile robot\footnote{\url{https://www.turtlebot.com/turtlebot3}}
to follow a $6$ meter path with $6$ waypoints, as shown in Figure \ref{fig: Turtlebot full path}.
We programmed the robot to perform the path by giving it a constant turn-rate of $0.2 [rad/s]$ for a direction and duration prescribed by the angle it should turn, followed by a constant forward velocity command of $0.1 [m/s]$ for a duration of 10-15 seconds for each segment of the path between two waypoints.
Using a motion-capture system\footnote{\url{https://optitrack.com/}}, we recorded $47$ repetitions of the robot completing the path unassisted.
For each of the path's waypoints, excluding the first and last, we provided the robot with relocation assistance by manually returning it to the correct position on the planned path $\pathplanned$ from which it continues execution.
We repeated this procedure for each of the 4 waypoints $27$ times, resulting in $155$ recorded cases of the robot completing the path with and without assistance.

As with the simulation, we paired the path with 1000 randomly generated cost-maps to calculate the AECD and VOA values for each of the 4 examined waypoints.

\noindent{\bf Results:}
For all but one waypoint on one map (i.e. $\frac{3999}{4000}$), VOA values fell within one standard deviation of AECD,  
demonstrating a strong correlation between AECD and \VOA. 

\begin{figure}[t!]
 \centering
\includegraphics[width=0.485\textwidth]{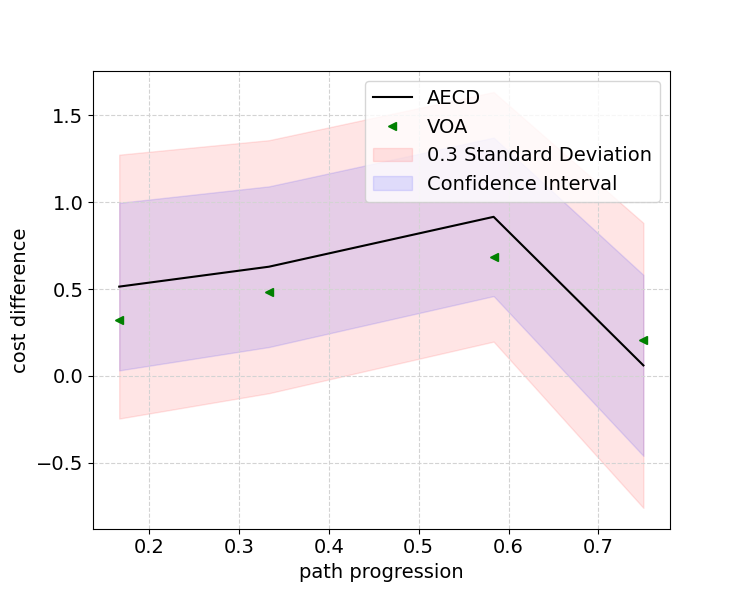}
 \caption{AECD and VOA (y-axis) per path progression $\locVar$ (x-axis) for a specific map. AECD is plotted with its 99\% confidence interval and 0.3 standard deviations.}
 \label{fig: real world resuls}
 \end{figure}

Figure \ref{fig: real world resuls} shows AECD and VOA per $\locVar$ for one map in this experiment (results for additional maps can be found in the supplementary material). 
Results show that the \VOA and AECD values are similar
to that obtained in the simulation, validating the results of the extensive simulations.
The key difference is that here we measured the actual movement of the robot instead of using an artificially generated noise kernel.

\subsection{Discussion}
Our results demonstrate how our \VOA~measures estimate the expected benefit of performing the considered assistive actions, even with large discrepancies between the modeled $\Delta \Sigma$ and the real covariance function of the agent's motion.

Our analysis promotes administering assistance to an agent at a waypoint where \VOA attains its maximal value. 
In order to use our measures to support the decision of which agent within a team should be assisted, there is a need to find a principled way to normalize the \VOA~values for the different agents in a way that is relevant to the specific application (e.g., according to path length).


Finally, we note that our framework can support any user-defined cost-map that reflects the objectives of the agent.
Specifically, for a setting in which an agent needs to follow a prescribed path, we can use a cost-map that assigns high rewards for the area around the waypoints on the path. A complete analysis of such a setting is given in the supplementary material.

  \section{Conclusion}

We introduced {\em Value of Assistance} (\VOA) for mobile agents and suggested ways to estimate \VOA~for two forms of assistance: relocation and localization. Our empirical evaluation demonstrated the ability of our \VOA~measures to predict the effect of the examined interventions in both simulated and real-world robotic settings. 

Future work will consider other forms of assistance (e.g., providing approximate locations) and other agent types (e.g., reward-maximizing replanning agents), and will account for optimization considerations of the helping agent (e.g., minimizing fuel consumption). 


\section*{Acknowledgements}
This research was supported by the Technion Autonomous System Program (TASP).

    \bibliographystyle{IEEEtran}
    
    \bibliography{bib}
    
\end{document}